\documentclass{article}
\usepackage{spconf,amsmath,graphicx}

\usepackage{enumitem}
\setlist{nosep, leftmargin=14pt}

\usepackage{mwe} % to get dummy images
\usepackage{hyperref}
\usepackage{amsmath}   % general math environments
\usepackage{amssymb}   % loads amsfonts and defines \mathbb
\usepackage{amsfonts}
\usepackage{booktabs}
\usepackage{multirow}
\usepackage{float}
\usepackage{xcolor}
\usepackage{setspace}

\newcommand{\eatme}[1]{ }

\title{A Multi-scale Linear-time Encoder for Whole-Slide Image Analysis}
\name{Jagan Mohan Reddy Dwarampudi{$^\dagger$} \quad Joshua Wong{$^\star$} \quad Hien Van Nguyen{$^\dagger$} \quad Tania Banerjee{$^{\ddagger\dagger}$}\vspace{-2mm}}
\address{$^{\dagger}$ Department of Electrical and Computer Engineering, University of Houston \\
       $^{\ddagger}$ Department of Information Science Technology, University of Houston \\
       $^{\star}$ Department of Neurology, College of Medicine, University of Florida}
\begin{document}
%\ninept
%
\maketitle

\begin{abstract}
We introduce Multi-scale Adaptive Recurrent Biomedical Linear-time Encoder (MARBLE), the first \textit{purely Mamba-based} multi-state multiple instance learning (MIL) framework for whole-slide image (WSI) analysis. MARBLE processes multiple magnification levels in parallel and integrates coarse-to-fine reasoning within a linear-time state-space model, efficiently capturing cross-scale dependencies with minimal parameter overhead. WSI analysis remains challenging due to gigapixel resolutions and hierarchical magnifications, while existing MIL methods typically operate at a single scale and transformer-based approaches suffer from quadratic attention costs. By coupling parallel multi-scale processing with linear-time sequence modeling, MARBLE provides a scalable and modular alternative to attention-based architectures. Experiments on five public datasets show improvements of up to \textbf{6.9\%} in AUC, \textbf{20.3\%} in accuracy, and \textbf{2.3\%} in C-index, establishing MARBLE as an efficient and generalizable framework for multi-scale WSI analysis.
\end{abstract}

\begin{keywords}
    Whole-slide image analysis, multiple instance learning, multi-scale modeling, state-space models, computational pathology
\end{keywords}
%
% \vspace{-5mm}
\section{Introduction}
\label{sec:intro}
\vspace{-3mm}
Multiple Instance Learning (MIL) is a core paradigm in computational pathology for gigapixel whole-slide images. A slide is represented as a \textit{bag} of patches with a slide-level label only. Early MIL used max or mean pooling, and later attention-based pooling highlighted salient regions but still treated patches as independent and identically distributed, limiting long-range spatial context \cite{ilse2018attention}. Transformer-based MIL introduced self-attention over patch embeddings \cite{shao2021transmil}, and hierarchical Vision Transformer pipelines such as HIPT preserved broader context with pyramids \cite{chen2022scaling}. Multi-scale MIL fused magnifications through hierarchical attention or semantic filtering (for example HAG-MIL and CS-MIL) \cite{xiong2023diagnose,deng2024cross}. In parallel, structured state-space models enabled efficient long-sequence modeling. The S4 model provided linear-time dynamics, and Mamba-based variants such as MambaMIL applied these ideas to pathology \cite{fillioux2023ssm_mil,yang2024mambamil}.

Despite this progress, efficient and scalable \emph{explicit} multi-scale reasoning remains lacking. First, many pipelines process a single magnification for stability and efficiency, leaving cross-scale dependencies unused. Second, transformer multi-scale fusion often incurs quadratic attention or multi-stage heuristics that hinder end-to-end training \cite{shao2021transmil,chen2022scaling}. Third, existing Mamba-based approaches typically treat each slide as a single-scale sequence, underusing the intrinsic multi-resolution pyramid. Taken together, these directions address efficiency, scalability, or context in isolation rather than in a unified linear-time framework.

We propose MARBLE, a state-space MIL framework that integrates coarse-to-fine reasoning within a linear-time backbone and \emph{processes multiple magnification levels in parallel via per-level modules} with lightweight cross-level fusion. Finer-level tokens are conditioned on their parent tokens from the coarser level, capturing cross-scale dependencies with minimal parameters and without quadratic attention. We use “level” as the primary term (synonymous with magnification/scale). Each level is handled by a state-space module with cost $\mathcal{O}(T_k)$ in its token count $T_k$, and parallel execution makes the wall-clock cost dominated by the finest level, yielding end-to-end behavior proportional to the largest $T_k$.

\begin{figure*}[t]
    \centering
    \includegraphics[width=0.87\textwidth]{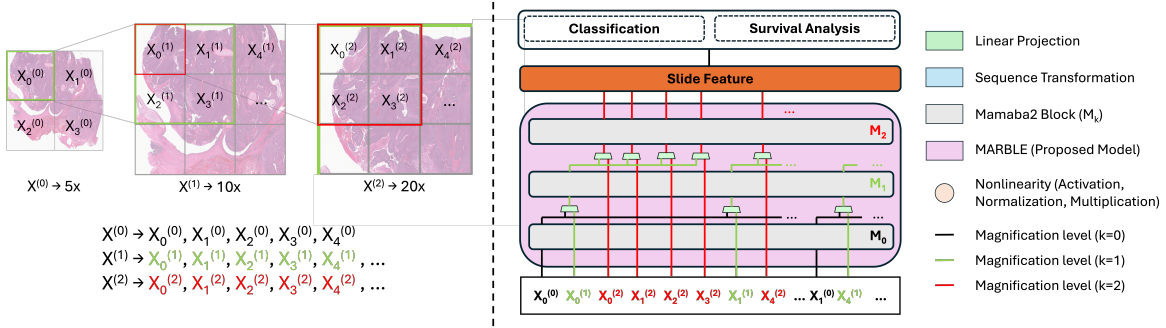}
    \vspace{-5mm}
    \caption{Overview of MARBLE. Left: a multi-resolution slide with three levels. Per-level sequences are formed and interleaved. Right: three Mamba-2 blocks \(M_k\) process levels \(k=0<1<2\). For \(k>0\), each token is fused with its parent from level \(k-1\). Final representations from \(M_2\) are pooled into a slide-level embedding.}
    \label{fig:sam_architecture}
    \vspace{-5mm}
\end{figure*}

The main contributions are:
\begin{enumerate}
    \item A fully Mamba-based MIL framework that introduces the first \emph{multi-scale encoding} with explicit coarse-to-fine fusion.
    \item A lightweight, token-aligned cross-level conditioning mechanism that injects coarse contextual information into finer representations, eliminating the need for quadratic attention or heavy fusion layers.
    \item Consistent gains across five public datasets for classification and survival, outperforming transformer- and Mamba-based baselines with minimal parameter overhead.
\end{enumerate}

\vspace{-5mm}
\section{Method}\label{sec:method}
\vspace{-3mm}
    % \subsection{Problem Setup and Notation}
    
        A whole-slide image (WSI) is provided at $S{+}1$ magnification levels indexed by $k\in\{0,1,\ldots,S\}$ with $k{=}0$ the coarsest and $k{=}S$ the finest. At level $k$, we tile the tissue region into non-overlapping $P{\times}P$ patches and extract $D$-dimensional embeddings
        $\mathbf{X}^{(k)}=\big[\mathbf{x}^{(k)}_{1}, \ldots, \mathbf{x}^{(k)}_{T_k}\big]^\top$, where $\mathbf{x}^{(k)}_{i}\in\mathbb{R}^{D}$ and $i\in\{1,\ldots,T_k\}$. Here $T_k$ denotes the number of tissue patches at level $k$, and we refer to each patch embedding $\mathbf{x}^{(k)}_i$ as a \emph{token}. We use the term \emph{level} to denote image magnification (or scale), where $k = 0$ corresponds to the lowest magnification (largest pixel size), and higher $k$ values indicate progressively finer resolutions.
        %We consistently use the term \emph{level}; ``magnification/scale'' are synonymous, with $k{=}0$ indicating lowest optical magnification (largest pixel size) and higher $k$ indicating finer resolution. 
        See Fig.~\ref{fig:sam_architecture} for an overview of levels and tokenization. We use bold uppercase for matrices/sequences and bold lowercase for vectors; subscripts index tokens; superscripts $(k)$ denote the level. % Indices standardized as $i\in\{1,\ldots,T_k\}$.
    \vspace{-4mm}
    \subsection{Multi-Scale Encoding with Coarse-to-Fine Fusion}
    \vspace{-2mm}
    Each level is encoded by an independent sequence module that operates in linear time with respect to $T_k$ (when using selective state-space models, the update equations follow the original backbone and are not repeated here; we refer readers to the source derivation~\cite{gu2024mamba}). Unless noted, each per-level encoder uses a single Mamba-2 block (depth $L{=}1$) with model dimension $D{=}1024$. We perform lightweight, token-aligned, coarse-to-fine fusion: \emph{before level $k{>}0$ is encoded}, each fine token is augmented with its parent context gathered from the \emph{output} at level $k{-}1$ (i.e., the encoded embeddings $\mathbf{Y}^{(k-1)}$),
    \vspace{-1mm}
    \[
    \mathbf{c}^{(k)}_{i} \;=\; \mathbf{y}^{(k-1)}_{p_k(i)}, 
    \qquad
    \tilde{\mathbf{x}}^{(k)}_{i} \;=\; \phi^{(k)}\!\big([\mathbf{x}^{(k)}_{i}\,\Vert\,\mathbf{c}^{(k)}_{i}]\big)\in\mathbb{R}^{D},
    \]
    \vspace{-1mm}
    where $\phi^{(k)}$ is a linear projection of token $\mathbf{x}^{(k)}_{i}$ concatenated with coarser parent token $\mathbf{c}^{(k)}_{i}$ with bias, and $\mathbf{Y}^{(k-1)}=\big[\mathbf{y}^{(k-1)}_{1},\ldots,\mathbf{y}^{(k-1)}_{T_{k-1}}\big]^\top$ are the token embeddings produced by the level-$k{-}1$ encoder. Parent–child mappings $p_k(i)$ are determined from the spatial tiling grid: patches at levels $k$ and $k{-}1$ are grid-aligned such that each fine tile falls entirely within exactly one coarse tile (Fig.~\ref{fig:sam_architecture}). In practice, $p_k(i)$ is obtained by integer division of the patch coordinates. If a fine-scale tile corresponds to a background region whose parent coarse tile was masked out, processing that fine-scale tile is also skipped.

Level-$k$ then encodes $\tilde{\mathbf{X}}^{(k)}=\big[\tilde{\mathbf{x}}^{(k)}_{1},\ldots,\tilde{\mathbf{x}}^{(k)}_{T_k}\big]^\top$ to produce $\mathbf{Y}^{(k)}$. All levels' execution wall-clock is dominated by the longest sequence (typically the finest level), while compute/memory scale linearly per level. This design preserves the linear-time complexity of each Mamba-2 encoder $\mathcal{O}(T_kD)$ while adding only minimal per-level projection parameters.

    \vspace{-4mm}
    \subsection{Slide-Level Heads}
    \vspace{-2mm}
    \noindent\textbf{Classification.}
    We first apply the coarse-to-fine fusion described above so that fine-level tokens are enriched with coarse context; \emph{only then} do we pool the finest-level representations using attention pooling:
    \vspace{-3mm}
    \[
    \begin{gathered}
    \mathcal{S}=\{\mathbf{y}^{(S)}_{i}\}_{i=1}^{T_S},\\[6pt] % add space here
    \alpha(\mathbf{y})=\frac{\exp(\mathbf{w}^{\top}\mathbf{y})}{\sum_{\mathbf{y}'\in\mathcal{S}}\exp(\mathbf{w}^{\top}\mathbf{y}')} ,\quad
    \mathbf{z}=\sum_{\mathbf{y}\in\mathcal{S}}\alpha(\mathbf{y})\,\mathbf{y}.
    \end{gathered}
    \]
    \vspace{-3mm}

    Here $\mathcal{S}$ is the set of encoded tokens at the finest level, $\mathbf{w}\in\mathbb{R}^{D}$ is the attention parameter, $\alpha(\cdot)$ are attention weights on $\mathcal{S}$, and $\mathbf{z}\in\mathbb{R}^{D}$ is the pooled slide representation. A linear classifier maps $\mathbf{z}$ to logits; we optimize cross-entropy.
    
    \noindent\textbf{Survival.}
    For cohorts with time-to-event labels, we attach a Cox proportional hazards head on $\mathbf{z}$ to produce a risk score $r\in\mathbb{R}$, with $r_i=\boldsymbol{\beta}^{\top}\mathbf{z}_i$, and minimize the negative partial log-likelihood
    \vspace{-1mm}
    \[
    \mathcal{L}_{\mathrm{Cox}} \;=\; - \sum_{i:\,\delta_i=1}\Big( r_i - \log \sum_{j\in \mathcal{R}(t_i)} e^{r_j} \Big) \;+\; \lambda \lVert\boldsymbol{\theta}\rVert_2^2,
    \]
    \vspace{-1mm}
    where $t_i$ is the event/censoring time, $\delta_i$ the event indicator, and $\mathcal{R}(t_i)$ the at-risk set (Breslow handling for ties). We apply an $\ell_2$ (ridge) penalty to the trainable parameters $\boldsymbol{\theta}$ via $\lambda\lVert\boldsymbol{\theta}\rVert_2^2$; $\boldsymbol{\beta}$ are the Cox weights.
    %$\boldsymbol{\beta}$ are the Cox weights and $\boldsymbol{\theta}$ collects trainable parameters regularized by $\ell_2$.
    \vspace{-4mm}
    \subsection{Regularization and Robustness}
    \vspace{-2mm}
    We apply two regularizers. \textbf{Random coarse-branch drop:} during training, randomly drop a fraction $\alpha$ of level-$0$ tokens and prune their descendants at all finer levels before fusion/encoding, creating stochastic sub-bags while preserving parent-child consistency. \textbf{Scan-order neutrality.} Within each level, the set of tokens (patch embeddings) is randomly permuted before encoding to prevent implicit positional bias. Because fusion depends solely on explicit parent lookups $p_k(i)$, MARBLE remains permutation-invariant with respect to the scan order at every level.

\vspace{-4mm}
\subsection{Datasets and Baselines}
\vspace{-2mm}
We evaluate MARBLE on two slide-level tasks. For diagnostic classification we use PANDA~\cite{bulten2022panda} with 5$\times$ and 20$\times$ scans and TCGA-NSCLC~\cite{tcga_program} with 10$\times$ and 40$\times$ scans. For survival analysis we use three TCGA cohorts with slide-level follow-up: KIRP, LUAD, and STAD. All slides are tiled into $256{\times}256$ patches. Both available magnifications are processed per dataset to match the parallel multi-level design (Fig.~\ref{fig:sam_architecture}). All WSI patch embeddings are extracted using the publicly released UNI model~\cite{chen2024uni} within the CLAM framework; weights are frozen during MARBLE training to isolate the effect of the multi-level encoder. Each patch is represented as a $D{=}1024$-dimensional vector. We then extract and process patches at both available magnifications for each slide, ensuring our analysis consistently employs two resolution levels across all datasets. 

\begin{figure}[ht]
    \vspace{-3mm}
    \centering
    \includegraphics[width=0.9\linewidth]{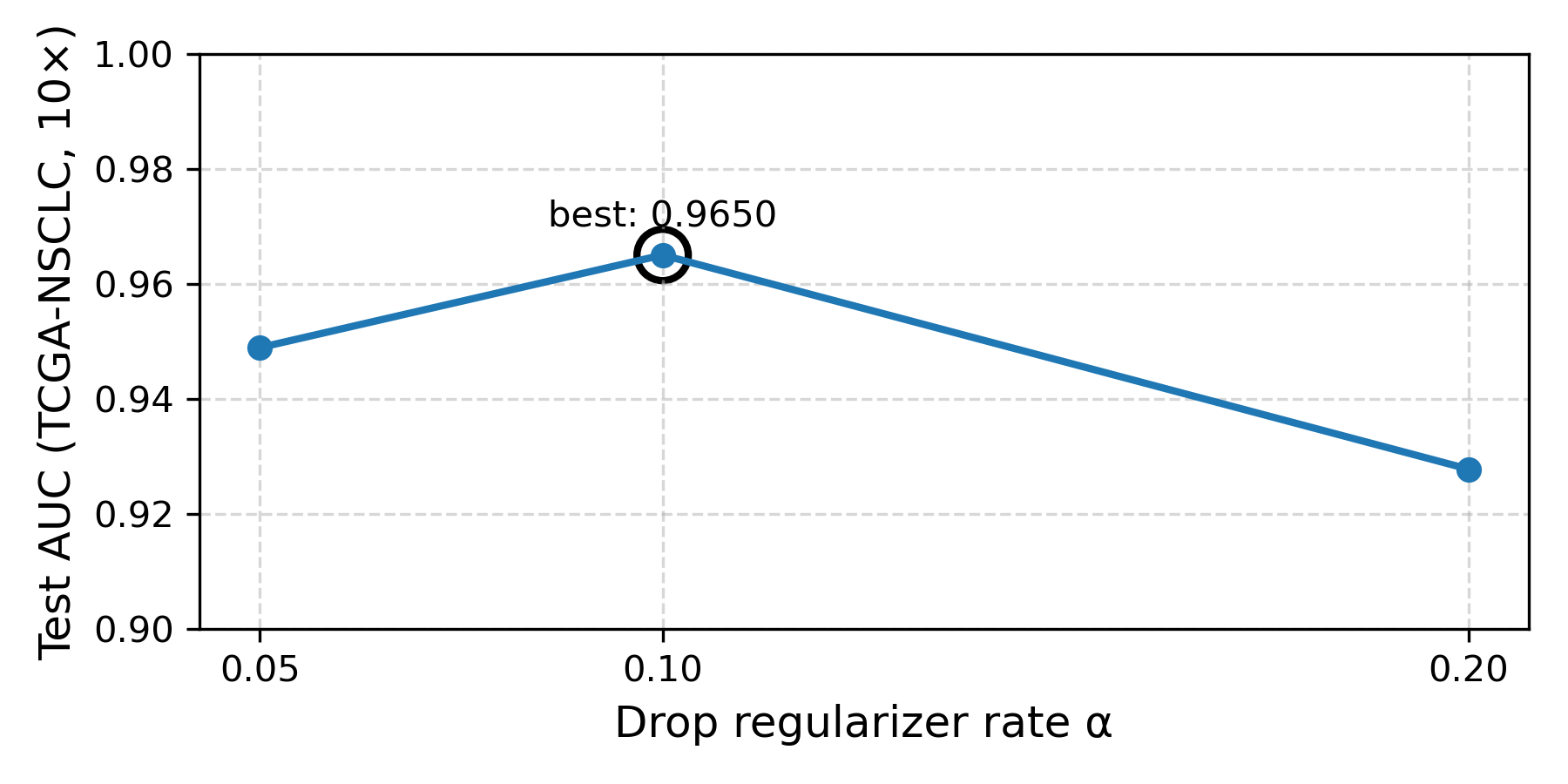}
    \vspace{-6mm}
    \caption{Effect of the drop regularizer rate $\alpha$ on test AUC for TCGA-NSCLC (10$\times$). $\alpha=0.1$ yields the best trade-off between regularization and retained discriminative signal.}
    \label{fig:drop_regularizer}
    \vspace{-3mm}
\end{figure}

We compare our model against a broad set of multiple instance learning approaches: ABMIL~\cite{ilse2018attention}, CLAM~\cite{lu2021data}, DSMIL~\cite{li2021dsmil}, TransMIL~\cite{shao2021transmil}, S4-MIL~\cite{fillioux2023ssm_mil}, DTFD-MIL~\cite{zhang2022dtfdmil}, MambaMIL~\cite{yang2024mambamil}, SRMambaMIL~\cite{yang2024mambamil}, and 2DMambaMIL~\cite{zhang2025_2dmamba}. Feature extraction and evaluation are kept identical across methods. Training uses AdamW~\cite{loshchilov2017decoupled} with $\beta_1{=}0.9$, $\beta_2{=}0.999$, weight decay $10^{-2}$, base learning rate $3{\times}10^{-5}$, cosine decay schedule with 5 warm-up epochs, and 30 total epochs. Early stopping was employed if the performance criteria validation AUC/C-index doesn't improve for consecutive 10 epochs. Slide-level batch size is $1$ due to variable token counts per slide and GPU memory limits. When official splits are unavailable we adopt an 80/10/10 train–val–test partition with a fixed seed. We report means over five repeated runs, using cross-validation or repeated splits as appropriate. Checkpoints are selected by validation AUC for classification and by C-index for survival. Reported metrics are expressed as mean across five runs. As shown in Fig~\ref{fig:drop_regularizer}, we also tune the drop regularizer via a small grid $\alpha\!\in\!\{0.05,0.1,0.2\}$ on a held-out split and fix $\alpha{=}0.1$ thereafter. All experiments were conducted on a single NVIDIA Tesla V100 GPU (32\,GB memory) using PyTorch. Code and pretrained checkpoints will be released %at \url{https://github.com/your-org/MARBLE-Mamba}.
upon publication.

\vspace{-4mm}
\section{Results}\label{sec:experiments}
\vspace{-3mm}

        \noindent\textbf{Classification.} MARBLE consistently outperforms strong MIL baselines. On PANDA, it surpasses the strongest competitor by \textbf{+20.25 percentage points (pp)} in accuracy and \textbf{+6.94 pp} in AUC. On TCGA-NSCLC, it improves over attention-based MIL and remains competitive with transformer and state-space variants. These gains align with the design choice to process magnifications in parallel and fuse them in a coarse-to-fine manner, which is especially helpful on larger bags where global context and high-resolution detail both matter. Full accuracy and AUC are in Table~\ref{tab:pruned_dataset_table}.

        \begin{table}[ht]
            \footnotesize
            \centering
            \renewcommand{\arraystretch}{1.2}
            \resizebox{0.9\columnwidth}{!}{%
                \begin{tabular}{l cc cc}
                    \toprule
                    
                    \multirow{2}{*}{\textbf{Method}} & 
                    \multicolumn{2}{c}{\textbf{PANDA}} & 
                    \multicolumn{2}{c}{\textbf{TCGA-NSCLC}} \\
                    
                    \cmidrule(lr){2-3} \cmidrule(lr){4-5}
                    
                    & \textit{Acc} & \textit{AUC}
                    & \textit{Acc} & \textit{AUC} \\
            
                    \midrule
                    
                    AB-MIL      & 0.4883 & 0.7797 & 0.8758 & 0.9572 \\
                    DSMIL       & 0.4633 & 0.7660 & 0.8782 & 0.9567 \\
                    CLAM        & 0.4802 & 0.7820 & 0.8804 & 0.9536 \\
                    DTFD-MIL    & 0.4704 & 0.7665 & 0.8736 & 0.9559 \\
                    TransMIL    & 0.4636 & 0.7728 & 0.8850 & 0.9626 \\
                    
                    \midrule
                    
                    S4-MIL      & 0.5047 & 0.7986 & 0.8851 & 0.9571 \\
                    MambaMIL    & 0.4679 & 0.7781 & 0.8758 & 0.9582 \\
                    SRMambaMIL  & 0.4711 & 0.7776 & 0.8850 & 0.9592 \\
                    2DMambaMIL  & 0.5075 & 0.8184 & 0.8851 & 0.9618 \\
            
                    \midrule
            
                    \textbf{MARBLE (Our Model)} & \textbf{0.7100} & \textbf{0.8878} & \textbf{0.8966} & \textbf{0.9730} \\
                    
                    \bottomrule
                \end{tabular}
            }
            \caption{Classification performance on two datasets. The highest metrics are \textbf{bold}.}
            \label{tab:pruned_dataset_table}
            \vspace{-3mm}
        \end{table}
        
        \noindent\textbf{Survival.} On KIRP, LUAD, and STAD, MARBLE attains the top C-index on all three cohorts. Improvements are most evident when both tissue architecture and cellular morphology drive prognosis. We report mean C-index over five repeats for each cohort in Table~\ref{tab:tcga_accuracies}.

        \begin{table}[htbp]
            \footnotesize
            \centering
            \resizebox{0.85\columnwidth}{!}{%
                \begin{tabular}{lccc}
                    \toprule
                    
                    \textbf{Method} & 
                    \textbf{KIRP} & 
                    \textbf{LUAD} & 
                    \textbf{STAD} \\
                    
                    \midrule
                    
                    ABMIL      & 0.7824 & 0.6157 & 0.6119 \\
                    DSMIL      & 0.7122 & 0.6114 & 0.6010 \\
                    CLAM       & 0.7197 & 0.5874 & 0.5883 \\
                    DTFD-MIL   & 0.7933 & 0.6020 & 0.6168 \\
                    TransMIL   & 0.7317 & 0.6139 & 0.5978 \\
                    
                    \midrule
                    
                    S4-MIL     & 0.7905 & 0.5945 & 0.6001 \\
                    MambaMIL   & 0.7822 & 0.5952 & 0.6244 \\
                    SRMambaMIL & 0.7424 & 0.5876 & 0.6130 \\
                    2DMambaMIL & 0.8027 & 0.6198 & 0.6428 \\
                    
                    \midrule
                    
                    \textbf{MARBLE (Our Model)} & \textbf{0.8184} & \textbf{0.6432} & \textbf{0.6510} \\
                    
                    \bottomrule
                \end{tabular}
            }
            \caption{Survival C-index comparison across three TCGA datasets. The highest metrics are \textbf{bold}.}
            \label{tab:tcga_accuracies}
            \vspace{-3mm}
        \end{table}

        \noindent\textbf{Ablation.} We compare single-scale models to the two-scale variant on TCGA-NSCLC (classification) and STAD (survival), evaluating \emph{coarse only} (10$\times$), \emph{fine only} (40$\times$), and \emph{combined} (with cross-scale fusion) under the same heads and training protocol. As shown in Table~\ref{tab:ablation_multiscale}, the combined setting is best on both tasks. It exceeds either single scale on TCGA-NSCLC in accuracy and AUC, and improves C-index on STAD over the stronger single-scale variant (10$\times$), indicating that coarse context stabilizes long-range reasoning while fine patches add discriminative local evidence.

        \begin{table}[htbp]
            \scriptsize
            \centering
            \resizebox{0.85\columnwidth}{!}{%
                \begin{tabular}{lccr}
                    \toprule
                    
                    \multirow{2}{*}{\textbf{Method}} & 
                    \multicolumn{2}{c}{\textbf{TCGA-NSCLC}} & 
                    \multicolumn{1}{c}{\textbf{STAD}} \\
                    
                    \cmidrule(lr){2-3}
                    \cmidrule(lr){4-4}
                    
                    & \textit{Acc} & \textit{AUC} & \textit{C-index} \\
                    
                    \midrule
                    
                    MARBLE (10$\times$)               & 0.8851 & 0.9650 & 0.6141 \\
                    MARBLE (40$\times$)               & 0.8736 & 0.9719 & 0.5961 \\
        
                    \midrule
                    
                    \textbf{MARBLE (10$\times$, 40$\times$)} & \textbf{0.8966} & \textbf{0.9730} & \textbf{0.6510} \\
                    
                    \bottomrule
                \end{tabular}%
            }
            \caption{Ablation results for multiple resolutions on TCGA-NSCLC (classification) and TCGA-STAD (survival). The highest metrics are \textbf{bold}.}
            \label{tab:ablation_multiscale}
            \vspace{-4mm}
        \end{table}
\vspace{-2mm}
\section{Conclusion}
\vspace{-1mm}

We presented MARBLE, a multiple‐instance learning framework that performs parallel multi-magnification encoding with lightweight coarse-to-fine fusion on a linear-time state-space backbone. By conditioning fine-scale tokens on coarse parents, MARBLE captures cross-scale dependencies without quadratic attention and minimal parameter overhead, with end-to-end cost dominated by the longest per-scale sequence. Across public datasets, MARBLE improves slide-level performance by up to \textbf{+20.3 pp} accuracy, \textbf{+6.9 pp} AUC, and \textbf{+2.3 pp} C-index. An ablation shows that combining coarse and fine magnifications consistently outperforms either alone, and a simple coarse-patch drop further regularizes training. Future work includes data-driven traversal, selective patch routing for efficiency, and evaluation with more than two native magnifications.

\vspace{-2mm}
\section{Compliance with Ethical Standards}
\vspace{-1mm}
This study was done retrospectively using human subject data made available in open access by the public datasets~\cite{ bulten2022panda, tcga_program}. Ethical approval was not required as confirmed by the licenses attached with the open access datasets.

\vspace{-2mm}
\section{Conflicts of Interest}
\vspace{-1mm}
This work was supported in part by the Center for Transformative Pathology and Health (CTPH) under UM1TR004539. The authors declare that they have no relevant financial or non-financial conflicts of interest.

%\bibliographystyle{IEEEbib}
%\bibliography{strings,refs}
\vspace{-2mm}
\def\IEEEbibitemsep{0pt plus .5pt}
\begin{spacing}{0.9}
\bibliographystyle{IEEEbib}
\bibliography{refs}

@inproceedings{ilse2018attention,
  title={Attention-based deep multiple instance learning},
  author={Ilse, Maximilian and Tomczak, Jakub and Welling, Max},
  booktitle={ICML},
  pages={2127--2136},
  year={2018},
}

@article{shao2021transmil,
  title={Transmil: Transformer based correlated multiple instance learning for whole slide image classification},
  author={Shao, Zhuchen and Bian, Hao and Chen, Yang and Wang, Yifeng and Zhang, Jian and Ji, Xiangyang and others},
  journal={NeurIPS},
  volume={34},
  pages={2136--2147},
  year={2021}
}

@inproceedings{loshchilov2017decoupled,
  title     = {Decoupled Weight Decay Regularization},
  author    = {Loshchilov, Ilya and Hutter, Frank},
  booktitle = {ICLR},
  year      = {2019},
}

@article{chen2024uni,
  title={Towards a General-Purpose Foundation Model for Computational Pathology},
  author={Chen, Richard J and Ding, Tong and Lu, Ming Y and Williamson, Drew FK and Jaume, Guillaume and Chen, Bowen and others},
  journal={Nature Medicine},
  publisher={Nature Publishing Group},
  year={2024}
}

@inproceedings{chen2022scaling,
  title={Scaling vision transformers to gigapixel images via hierarchical self-supervised learning},
  author={Chen, Richard J and Chen, Chengkuan and Li, Yicong and Chen, Tiffany Y and Trister, Andrew D and Krishnan, Rahul G and Mahmood, Faisal},
  booktitle={CVPR},
  pages={16144--16155},
  year={2022}
}

@article{gu2024mamba,
  title={Transformers are ssms: Generalized models and efficient algorithms through structured state space duality},
  author={Dao, Tri and Gu, Albert},
  journal={arXiv preprint arXiv:2405.21060},
  year={2024}
}

@inproceedings{yang2024mambamil,
  title={Mambamil: Enhancing long sequence modeling with sequence reordering in computational pathology},
  author={Yang, Shu and Wang, Yihui and Chen, Hao},
  booktitle={MICCAI},
  pages={296--306},
  year={2024},
  organization={Springer}
}

@article{deng2024cross,
  title={Cross-scale multi-instance learning for pathological image diagnosis},
  author={Deng, Ruining and Cui, Can and Remedios, Lucas W. and Bao, Shunxing and Womick, R. and Chiron, Sophie and others},
  journal={Medical Image Analysis},
  volume={94},
  pages={103124},
  year={2024}
}

@inproceedings{xiong2023diagnose,
  title={Diagnose like a pathologist: Transformer-enabled hierarchical attention-guided multiple instance learning for whole slide image classification},
  author={Xiong, Conghao and Chen, Hao and Sung, Joseph J. Y. and King, Irwin},
  booktitle={IJCAI},
  pages={1596--1604},
  year={2023}
}

@inproceedings{fillioux2023ssm_mil,
  title={Structured state space models for multiple instance learning in digital pathology},
  author={Fillioux, L\'eo and Boyd, Joseph and Vakalopoulou, Maria and Courn\`ede, Paul-Henry and Christodoulidis, Stergios},
  booktitle={MICCAI},
  pages={594--604},
  year={2023}
}

@misc{tcga_program,
  author={{TCGA Research Network}},
  title={The Cancer Genome Atlas (TCGA) Program},
  year={2016}
}

@article{bulten2022panda,
  title={Artificial intelligence for diagnosis and Gleason grading of prostate cancer: the PANDA challenge},
  author={Bulten, Wouter and Kartasalo, Kimmo and Chen, Po-Hsuan Cameron and others},
  journal={Nature Medicine},
  volume={28},
  number={1},
  year={2022}
}

@inproceedings{li2021dsmil,
  title={Dual-stream multiple instance learning network for whole slide image classification},
  author={Li, Bin and Li, Yao and Eliceiri, Kevin W.},
  booktitle={CVPR},
  year={2021}
}

@inproceedings{zhang2022dtfdmil,
  title={DTFD-MIL: Double-tier feature distillation multiple instance learning for histopathology whole slide image classification},
  author={Zhang, Hongrun and Meng, Yanda and Zhao, Yitian and Qiao, Yihong and Yang, Xiaoyun and Coupland, Sarah E. and Zheng, Yalin},
  booktitle={CVPR},
  year={2022}
}

@inproceedings{zhang2025_2dmamba,
  title={2DMamba: Efficient State Space Model for Image Representation with Applications on Giga-Pixel Whole Slide Image Classification},
  author={Zhang, Jieru and others},
  booktitle={CVPR},
  year={2025}
}

@article{lu2021data,
  title={Data-efficient and weakly supervised computational pathology on whole-slide images},
  author={Lu, Ming Y and Williamson, Drew FK and Chen, Tiffany Y and Chen, Richard J and Barbieri, Matteo and Mahmood, Faisal},
  journal={Nature biomedical engineering},
  volume={5},
  number={6},
  pages={555--570},
  year={2021},
  publisher={Nature Publishing Group UK London}
}
\end{spacing}

\end{document}